\documentclass{article}
\PassOptionsToPackage{numbers, compress}{natbib}

\usepackage[preprint]{neurips_2022}
\usepackage[utf8]{inputenc} % allow utf-8 input
\usepackage[T1]{fontenc}    % use 8-bit T1 fonts
\usepackage{hyperref}       % hyperlinks
\usepackage{url}            % simple URL typesetting
\usepackage{booktabs}       % professional-quality tables
\usepackage{amsfonts}       % blackboard math symbols
\usepackage{amsmath}
\usepackage{amssymb}
\usepackage{amsthm}
\usepackage{nicefrac}       % compact symbols for 1/2, etc.
\usepackage{microtype}      % microtypography
\usepackage{xcolor}         % colors
\usepackage{graphicx}		% for setting images
\usepackage{multirow}
\usepackage{subcaption}

\title{Neural Additive Models for Nowcasting}

\author{%
  Wonkeun Jo
    \\
  Department of Computer Science\\
  Chungnam National University\\
  99 Daehak-ro, Yuseong-gu, Daejeon \\
  \texttt{jowonkun@o.cnu.ac.kr} \\
  \And
  Dongil Kim\thanks{\ indicates the corresponding author.} \\
  Department of Computer Science\\
  Chungnam National University\\
  99 Daehak-ro, Yuseong-gu, Daejeon \\
  \texttt{dkim@cnu.ac.kr} \\
}

\begin{document}

\bibliographystyle{NIPS_bst} % NIPS format

\maketitle

\begin{abstract}
Deep neural networks (DNNs) are one of the most highlighted methods in machine learning. However, as DNNs are black-box models, they lack explanatory power for their predictions. Recently, neural additive models (NAMs) have been proposed to provide this power while maintaining high prediction performance. In this paper, we propose a novel NAM approach for multivariate nowcasting (NC) problems, which comprise an important focus area of machine learning. For the multivariate time-series data used in NC problems, explanations should be considered for every input value to the variables at distinguishable time steps. By employing generalized additive models, the proposed NAM-NC successfully explains each input value’s importance for multiple variables and time steps. Experimental results involving a toy example and two real-world datasets show that the NAM-NC predicts multivariate time-series data as accurately as state-of-the-art neural networks, while also providing the explanatory importance of each input value. We also examine parameter-sharing networks using NAM-NC to decrease their complexity, and NAM-MC’s hard-tied feature net extracted explanations with good performance.
\end{abstract}

\section{Introduction}\label{sec.1}

Deep neural networks (DNNs) are among the most powerful machine learning tools used for natural language processing \cite{electra}, image classification \cite{yolov3}, sound recognition \cite{TCN}, and more. DNNs are powerful because of their free input and output forms, which other methods lack \cite{adv_nn}. This advantage can be applied to real-world time-series data problems, such as nowcasting (NC), which uses a DNN to predict the next time step for power system monitoring \cite{Informer}, rainfall prediction \cite{rainfall}, cutting-force prediction, and other problems. 

Because DNNs are black-box models, they lack explanatory power behind their predictions. This is a critical issue, as the scientific method requires a good understanding of the system and the causes of the results. Neural additive models (NAMs) have recently been proposed to help explain how DNNs reach their prediction values \cite{nam}. NAMs assign targets to feature-net base modules that represent each variable. Then, based on their changes, the reasons for the DNN output are better understood. 

In this paper, we propose a novel NAM for multivariate NC problems (NAM-NC) by extending them to explain the predictions of multivariate time-series data. An explanation must be considered for each input value of the variables at all time steps. By employing a generalized additive model (GAM), the proposed NAM-NC successfully explains each input value’s importance. Furthermore, it supports multi-task learning and is suitable for multivariate NC tasks. According to our results, the feature-nets maintain prediction performance, even when the NAMs are configured with fewer than optimal parameters.

This work makes the following contributions:

\begin{itemize}
    \item GAMs enable NAM-NC to explain NC results after training without additional model-agnostic methods \cite{gam}.
    \item Multivariate NC is supported, and each NC is influenced by cross-correlating other time series, including auto-correlations.
    \item Two-parameter sharing structures reduce NAM-NC’s complexity, and with reduced parameters, its performance is maintained.
    \item State-of-the-art performance is maintained for time-series NC tasks using real datasets.
\end{itemize}

We review past works in Section 2. Section 3 introduces NAM-NC and illustrates how it explains the prediction mechanisms of DNNs using synthetic sinusoidal signals. Section 4 explains how to share parameters in NAM-NC. Section 5 uses experimentation to demonstrate the competitiveness of NAM-NC with other time-series DNN methods on a benchmark dataset. This section also visualizes NAM-NC's calculation methods. Finally, Section 6 concludes this work.

\section{Related work}\label{sec.2}

The use of auto-regressive (AR) models to tackle time-series NC problems has long been a key focus of machine learning. 
The AR model predicts the value of the next time step by calculating the weights that best explain the data in the previous step \cite{ar_theory}. 
The vector AR method was designed for multivariate NC by extending the univariate NC method \cite{vectorAR}. 
Owing to their simplicity, basic AR models can provide predictive interpretations of derivative models that are still in use today \cite{ar_nowcasting}. 
A recent AR project, Prophet, performed univariate NC tasks \cite{prophet}. 
There have also been attempts to estimate AR parameters using NN structures \cite{ARnet, arconv}. 
Prophet was later improved to learn traditional statistical solutions this way \cite{neural_prophet}.
% , but time-series NC data are often processed from datasets using tabular formats \cite{gbmnowcast1}.
Also, time-series NC data are often processed from datasets using non-recursive tabular formats \cite{gbmnowcast1}.

New NN methods have been introduced to extract the most meaningful vectors from time series. 
Long short-term memory (LSTM) improves recurrent NNs (RNNs), enabling them to retain meanings extracted from more distant time points \cite{lstm}. 
Other studies have attempted to compensate for the shortcomings of LSTM \cite{lstma,hst-lstm,lstnet}. 
Meanwhile, attention-based and convolutional NNs (CNNs) have been applied to time-series predictions, showing good performance \cite{attention,cnn}. 
The bidirectional-encoder-representations-from-transformers (BERT) model with an attention layer is among the most powerful natural language processing (NLP) tools used today \cite{bert}. 
Notably, the positional encoding used by BERT to convey location information in the attention layer is effective for time-series learning \cite{positionale}. 
The Informer application achieved state-of-the-art (SOTA) performance with time-series forecasting by reducing the computational costs of the attention layer and improving positional encoding \cite{Informer}. 
A temporal CNN using dilated convolution is a representative model that processes time-series based on a CNN \cite{dilated, TCN}. 
Recently, SCINet achieved SOTA on benchmark datasets by convolving a binary tree-like hierarchical structure \cite{scinet}.

However, the explainability power of DNN prediction mechanisms has been elusive \cite{exai2}. 
Thus, researchers have shown interest in correlating DNN predictions to the variability of machine learning input \cite{exai1}. 
Class activation mapping estimates the contributions of input data by relying on the CNN’s receptive field \cite{cam}. 
The temporal LSTM with attention also interprets the time series by employing an attention layer to an RNN structure \cite{exrnn}. 
Layer-wise relevance propagation suggests the possibility of interpreting DNN results by backtracking the values contributing to the prediction results in the NN graph \cite{lrp}. 
The local-interpretable-model-agnostic-explanations model interprets a prediction using hypothetical data close to the predicted data \cite{lime}. 
Meanwhile, GAMs make predictions using contributions extracted by a simple function for each variable in the data \cite{gam}. 
The GAM function is presented in Eq. \ref{eq:gam}. The GAM uses the base module $f$ that only focuses on one of the $K$ variables for explainable predicting.
Because the GAM employs an isolated module $f$ for each variable, the GAM provides the influence of each variable for predicting the target $\hat{y}$.

\begin{align}
\begin{split}\label{eq:gam}
    \hat{y} =& \beta + f_1(x_1) + \cdots + f_K(x_K) = \beta + \sum_{k = 1}^{K} f_k(x_k)
\end{split}
\end{align}

The Shapley additive explanations (SHAP) model describes popular NN methods based on GAM and Shapley theory \cite{shap}.
In particular, SHAP shows that DNNs can be described as model-agnostic using DeepLift, which compares the activation of each neuron to its reference activation and assigns contribution scores according to the difference \cite{deeplift}. The time-SHAP model further explains LSTMs and reduces the computational cost of kernel-SHAP during explanatory training \cite{timeshap}.

\section{NAM-NC}\label{sec.3}

NAM extracts features from each variable using a simple feature net. 
Contrary to recent trends, NAM has produced good results with tabular data, despite being a lightweight network \cite{nam,deit}.
Thus, we employed the basic premise of NAM and expanded it to handle NC tasks by employing multi-task learning to support predictions based on multivariate time-series data.

\begin{figure}[!t]
\centering
\includegraphics[width = .75\textwidth]{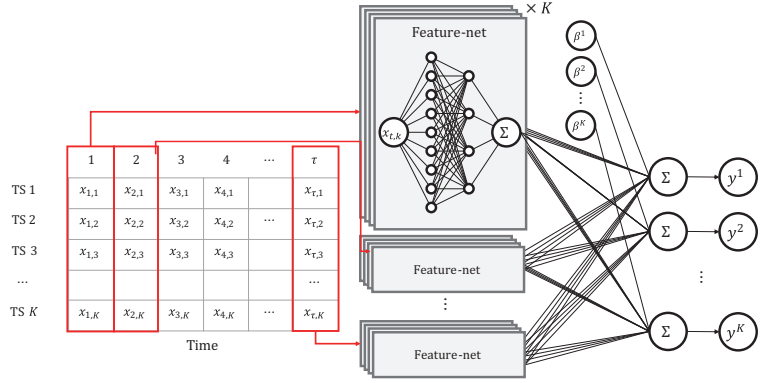}
\caption{Neural additive model for nowcasting.}
\label{fig:arch}
\end{figure}

Figure \ref{fig:arch} illustrates the structure of NAM-NC for multivariate time-series NC. 
As shown, the time series, $X_{T-\tau+1:T,K}\in\mathbb{R}^{\tau \times K}$, can be expressed in the form of $K$ variables and with $\tau$ length.
Each value, $x_{t,k}$, of the time series, $X$, is input to the feature-net, $f_{t,k}$.
Each feature net receives one variable, $x_k\in\mathbb{R}^{1}$, and computes one feature scalar, $f_k(x_k)\in\mathbb{R}^{1}$, with the same dimension.
The goal of NC is to predict $X_{T+1,K} = Y$ at time $T+1$ based on $X_{T-\tau+1:T,K}$.
We can predict the $T+1^{th}$ time step of the $k^{th}$ series by again calculating a linear combination of the extracted feature values and adding them.
Thus, Eq. \ref{eq:nam_nc} expresses the NC result of Eq. \ref{eq:gam}:

\begin{align}
\begin{split}\label{eq:nam_nc}
    \hat{y}^K = \beta^K + w^K_{1,1}f_{1,1}(x_{1,1}) + \cdots + w^K_{\tau,K}f_{\tau,K}(x_{\tau,K}) = \beta^K + \sum_{t = 1}^{\tau} \sum_{k = 1}^{K} w^K_{t,k}f_{t,k}(x_{t,k}).
\end{split}
\end{align}

The superscripts of all items in the Eq. \ref{eq:nam_nc} specify time-series $K$ among those located at time $T+1$.
Eq. \ref{eq:nam_nc} provides the process of summing the feature values extracted by the feature net and weighting them according to their location information. The scalar extracted by the feature-net, $f_{t,k}(\cdot)$, is influenced only by the input value, $x_{t,k}$. 
As the extracted scalar does not combine with other time-series points during training, $f(x)$ is used as the specific input value. Especially, the NAM utilizes the Exp-centered Unit (ExU) as its feature net $f$, which is designed to be sensitive to the contribution of the input \cite{nam}.
Therefore, NAM-NC can ignore unrelated time-series points. 
However, these important points can greatly affect prediction outcomes.

\begin{figure}[!t]
\centering
\includegraphics[width = .9\textwidth]{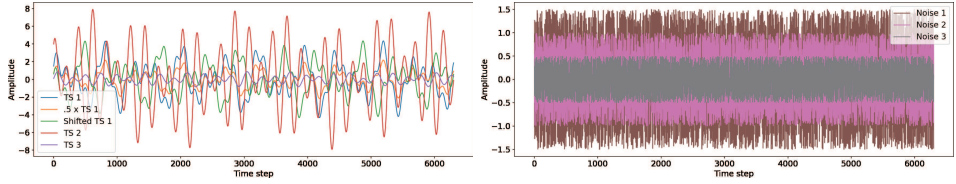}
\caption{Five synthesized time series and three noises.}
\label{fig:syn_sample}
\end{figure}

Next, we provide an analysis of NAM-NC results using simple synthetic time-series toy data consisting of five relevant features and three irrelevant random noise items. 
The relevant features are generative and combine the sine and cosine functions of different periods and amplitudes. 
Figure \ref{fig:syn_sample} shows the generated time-series data. 
One of the base time series, TS1, is used as the original, from which we produce two variations. 
The other time series, TS2 and TS3, are not related. 
One variation, ``.5 x TS 1," only halves the amplitude while retaining the TS1's period. 
The other, ``Shifted TS 1," employs a periodic shift with the same amplitude as the original.
For the experiments, we set $\tau$ to eight for all time series. 
We applied NAM-NC to the multivariate time-series data, expecting that the relevant features would provide high feature-net outputs and that the irrelevant ones would provide low outputs.

\begin{figure}
\centering
\includegraphics[width = .9\textwidth]{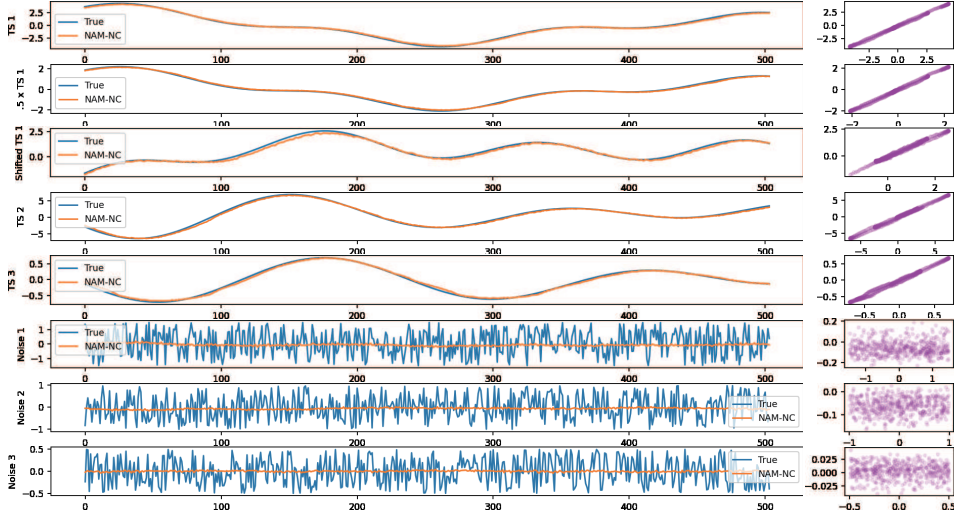}
\caption{Nowcasting performance for eight time series.}
\label{fig:syn_train}
\end{figure}

\begin{figure*}[!t]
     \centering
     \begin{subfigure}[b]{.45\textwidth}
         \centering
         \includegraphics[width=\textwidth]{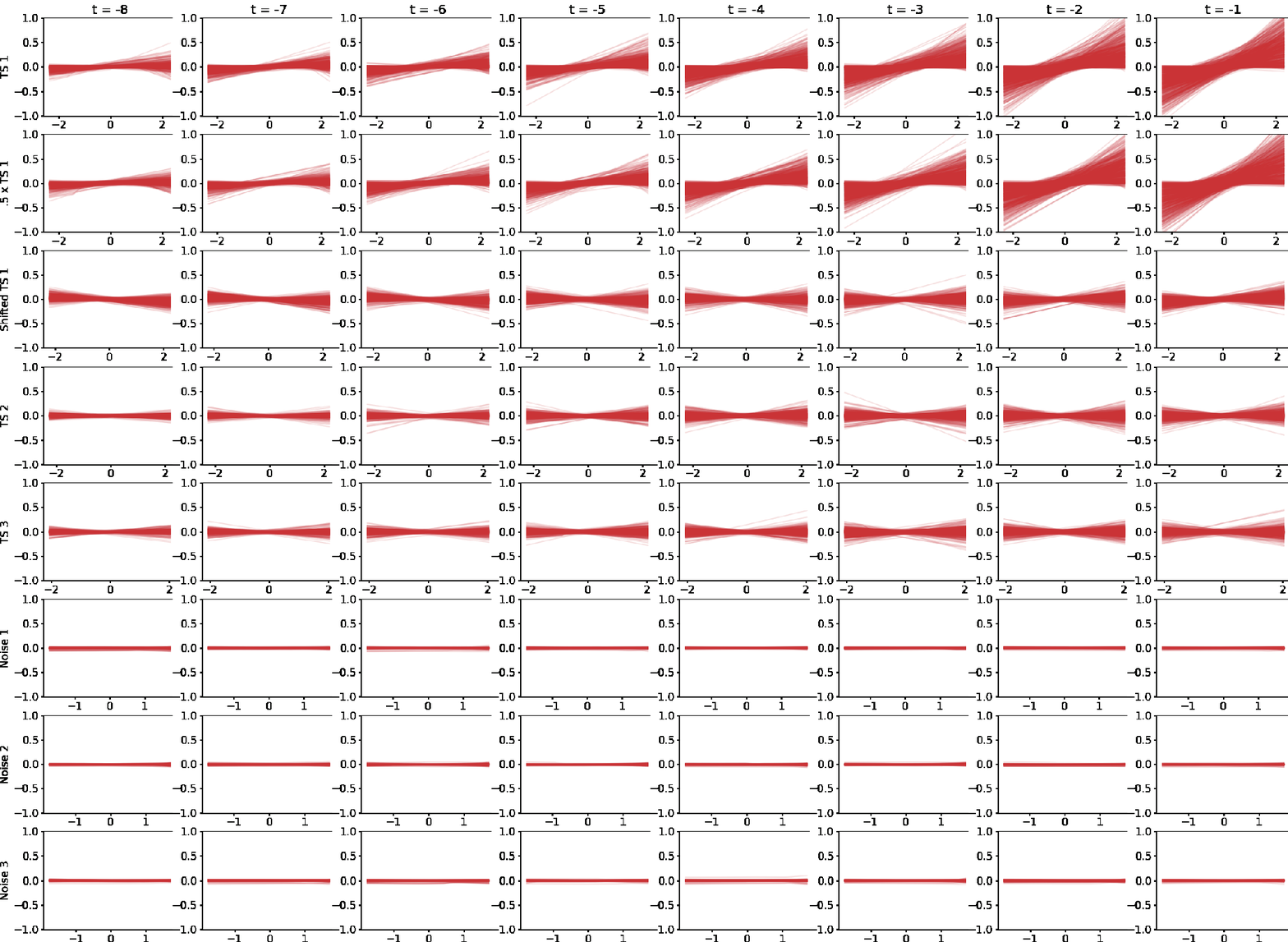}
     \end{subfigure}
     \hfill
     \begin{subfigure}[b]{.45\textwidth}
         \centering
         \includegraphics[width=\textwidth]{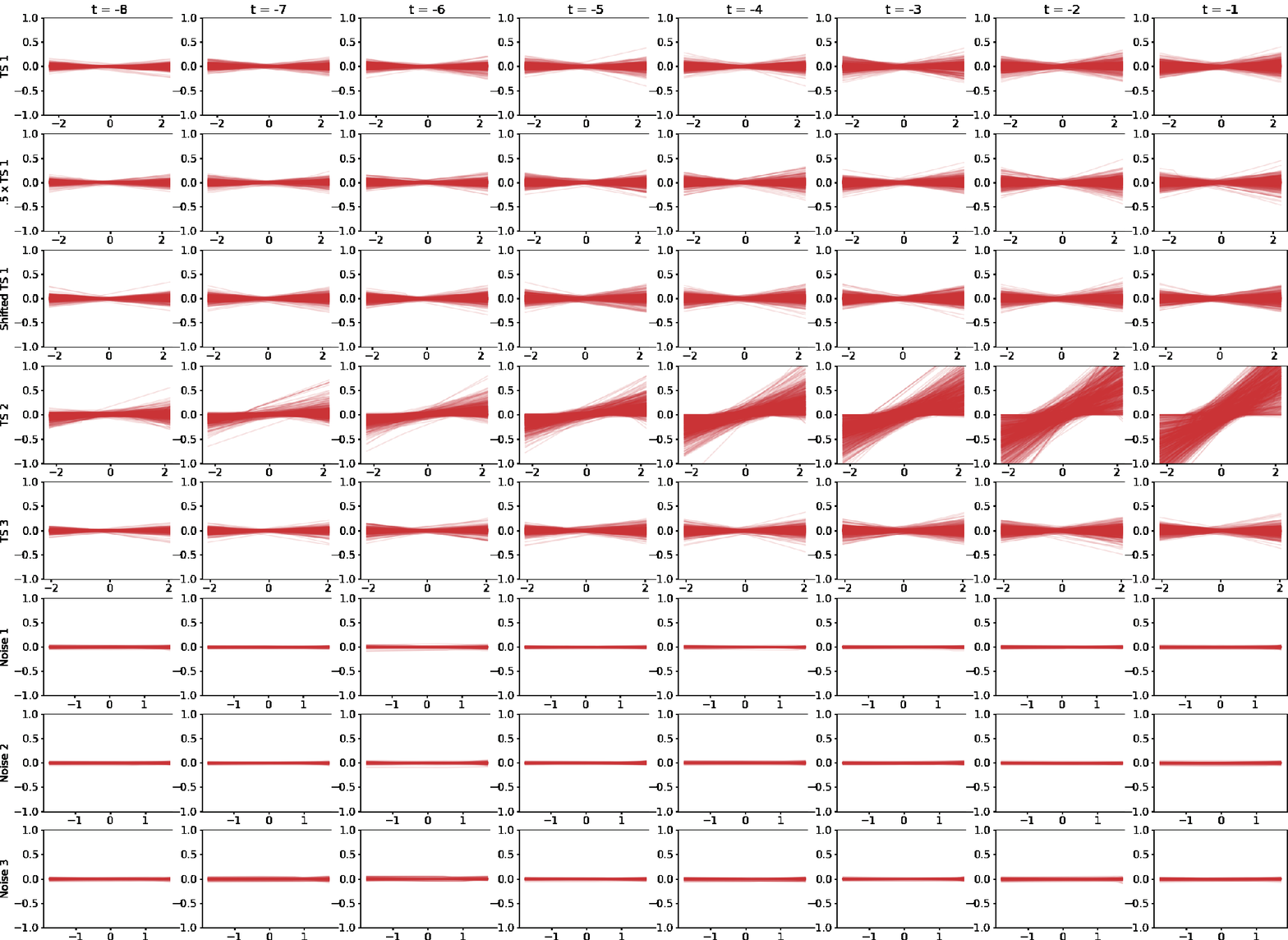}
     \end{subfigure}
        \caption{Contribution according to the unique value of each time series. The nowcasting target of left and right figure is the TS1 and TS2, respectively.}
        \label{fig:syn_cont}
\end{figure*}

Figure \ref{fig:syn_train} illustrates the NC results of the synthesized time-series data. 
The sub-figures in the first column indicate each true point in the time series (blue line) and the predicted points of NAM-NC (orange line) in the time step domain. 
The sub-figures in the second column express the same information as an x- vs. y-axis slope. 
Thus, it is apparent that NAM-NC can accurately predict multivariate time-series data. 
Figure \ref{fig:syn_cont} illustrates the importance of each time-series point to prediction.
NAM-NC employs $\tau \times K$ feature nets to extract the importance from scalar data points for each. 
Figure \ref{fig:syn_cont} illustrates the contributions of $\tau \times K$ input points to TS1 (left) and TS2 (right). 
Rows and columns indicate each $k$ time series and $t$ step, respectively. 
The x-axis of the subplots located in each matrix represents the unique values of the input time series, and the y-axis is the value extracted by $f$ dedicated to each point from the unique values of the input.
We repeated training 1,000 times using the same structure to check the consistency of NAM-NC. 
Figure \ref{fig:syn_cont} presents all contributions made by passing the unique $k^{th}$ time series through 1,000 times repeat. 
The red line is translucent to make it easier to see that training was repeated. 
According to the results, We figured out that the feature net $f_{t,k}$ of NAM-NC nowcasting the time series $k$ produces a smaller contribution as it looks at the distant past time.
In TS1’s NC, only the feature scalars extracted from the two time series related to TS1 made significant contributions. 
Moreover, TS2’s NC only obtained its contributions from the time series to which it belonged. 
We also found that no NC targets received contributions from noises.

% prof. ver
% In addition, figure \ref{fig:syn_cont} illustrates the importance of each time series points for prediction.
% NAM-NC's nowcasting employs $\tau \times K$ scalar data points for each prediction.
% Figure \ref{fig:syn_cont} illustrates the contributions of $\tau \times K$ input points to TS1 (left) and TS2 (right).
% Row and column indicate each time series and time step, respectively.
% The X-axis of the subplots located in each matrix represents the unique values of the input time series.
% And the Y-axis of the subplot is the value extracted by $f$ dedicated to each point from the unique values of the input time series.
% In particular, we repeated the training 1,000 times of NAM-NC respectively with the same structure to check the consistency of NAM-NC.
% And figure \ref{fig:syn_cont} represents all the contributions that occurred when the unique data in training was passed to 1000 different NAM-NCs.
% So, the red line in the figure is translucent to make it easier to see that training was repeated.
% According to the experiment, in nowcasting TS1, only the feature scalars extracted from the two time series related to TS1 make a large contribution.
% Moreover, the nowcasting for TS2 only gets its contribution from the time series it belongs.
% Simultaneously, we find that all nowcasting targets don't receive any contribution from the noise.

\section{NAM-NC parameter sharing}\label{sec.4}

One advantage of NAM is that it is possible to find the sufficient context vector needed for inference from table data, even when using a simple unit network \cite{nam}.
NAM-NC repeats the number of variables as many times as the value of time-step $\tau$, resulting in a higher computational cost than regular NAM. 
Therefore, we sought a way to reduce the computational cost of NAM-NC. 
A simple potential solution is parameter sharing, which in multi-task learning is often used to reduce the weight of NNs \cite{mtl_param_share1}. 
Hard-sharing, which involves sharing most parameters while independently utilizing only the output layer for multi-task learning, can also significantly reduce weight \cite{mtl_param_share2}. 
Thus, we applied hard-sharing in terms of time or features between feature-nets to tie the parameters of NAM-NC. 
Its formula, as applied to hard-sharing, is shown in the two-part Eq. \ref{eq:nam_nc_time}:

\begin{align}
\begin{split}\label{eq:nam_nc_time}
    \hat{y}^K_{time} =& \beta^K + w^K_{1,1}f_{:,1}(x_{1,1}) + \cdots + w^K_{\tau,K}f_{:,K}(x_{\tau,K}) = \beta^K + \sum_{t = 1}^{\tau} \sum_{k = 1}^{K} w^K_{t,k}f_{:,k}(x_{t,k}) \\ 
    \hat{y}^K_{feature} =& \beta^K + w^K_{1,1}f_{1,:}(x_{1,1}) + \cdots + w^K_{\tau,K}f_{\tau,:}(x_{\tau,K}) = \beta^K + \sum_{t = 1}^{\tau} \sum_{k = 1}^{K} w^K_{t,k}f_{t,:}(x_{t,k})
\end{split}
\end{align}

Where “:” refers to the dimension with which the parameter is shared. Accordingly, we can reduce the computational cost by sharing feature-net $f$ at each time step or series to increase the computational cost of NAM-NC. Therefore, the space complexity of NAM-NC's parameters is reduced from $\mathcal{O}(\tau \times K)$ to $\mathcal{O}(\tau)$, or $\mathcal{O}(K)$.

% 반복문 관련한 내용 추가
\section{Evaluating the performance of the NAM-NC}\label{sec.5}

Here, we compare the NC performance of NAM-NCs to that of other SOTA models using real data.
All our experiments are available on the NAM-NC GitHub site. 
For model comparison, we prepared Informer\footnote{https://github.com/zhouhaoyi/Informer2020}, for which temporal encoding is applied with positional encoding, and SCINet\footnote{https://github.com/cure-lab/SCINet} and LSTM, for which both had the input format required.
Temporal encoding was used with Informer to capture the real-time influence of variables.
However, as we did not use real-time information for NC, we removed its temporal encoding and kept the positional encoding. 
SCINet performs convolution by splitting the time series into a binary tree structure.
Thus, the length of its input time series must take the format of $2^n$. 
Therefore, we used a time-series input length of eight for all models.

% \usepackage{multirow}
% \usepackage{graphicx}
\begin{table}[!t]
\caption{Configuration of the experiment.}
\label{tab:ex_config}
\vspace{10pt}
\centering
\resizebox{.80\textwidth}{!}{%
\begin{tabular}{lrclrc}\toprule
Model name    & Param. name    & Set      & \multicolumn{2}{r}{Param. name}  & Set   \\ \midrule
\multirow{2}{*}{Shared params} & Batch-size             & 128                  & \multicolumn{2}{r}{Early stop round}  & 10   \\
                              & lr                & 0.001                & \multicolumn{2}{r}{Dropout}             & 0.1  \\\midrule
\multirow{3}{*}{NAM-NC}        & Exu –- linears      & {[}100,32{]}         & \multirow{3}{*}{LSTM}   & hidden        & 100  \\
                              & Activation             & leaky--ReLU           &                         & n--layer      & 2    \\
                              & Init distribution & $\mathcal{N}(0,1)$ &                         & bidirectional & True \\\midrule
\multirow{3}{*}{Informer}      & $d_{model}$          & 64                   & \multirow{3}{*}{SCINet} & Num levels   & 3    \\
                              & n--heads          & 4                    &                         & Kernel size  & 5    \\
                              & $d_{ff}$             & 64                   &                         & Conv hidden        & 1   \\\bottomrule
\end{tabular}%
}
\end{table}

Table \ref{tab:ex_config} summarizes the hyperparameters used for the experiment. 
The top row identifies the categories of the other rows. 
Specific parameters are shown for each model. 
Unimportant parameters were set to their default values according to their respective research articles. 
We employed the Adam optimizer with no weight decay. 
The feature net of NAM-NC first passes through ExU and then through its linear combination layer \cite{nam}. 
According to NAM, 1,000 nodes of ExU are sufficient to extract the feature scalars from all inputs.
However, in our experiment, only 100 nodes of ExU were used.
Thus, we extracted the feature scalar from the TS points without using a linear combination multilayer structure. 
Additionally, a rectified linear unit (ReLU) with a limited maximum value for NAM was used to lower the dependence of specific feature-nets while hindering training the time-series data. 
Thus, we used a leaky ReLU for NAM-NC’s feature-net activation.

\begin{figure*}[!b]
\centering
\includegraphics[width = .7\textwidth]{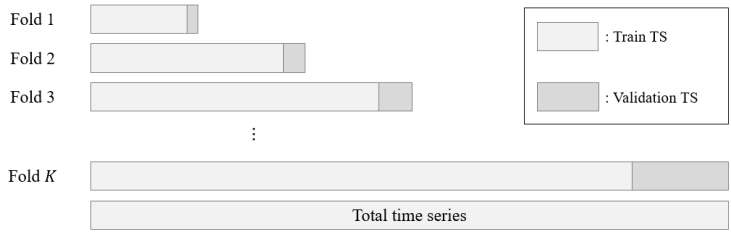}
\caption{Process of the time series $K$--fold cross-validation.}
\label{fig:TS_kfold_ex}
\end{figure*}

During the time-series cross-validation process, after dividing the time series into $K$ equal parts, the training time series was accumulated from the front, followed by $K$ instances \cite{ts_kfold}.
The subsequent time series for validation was cropped by 10\% of its length and used for learning. 
Figure \ref{fig:TS_kfold_ex} illustrates the cross-validation process. 
Next, we set $K$ to 10 for NC validation.
We then used the root mean-squared error (RMSE), the mean absolute error (MAE), and $R^2$ as criteria for validation.
RMSE and MAE reflect the absolute deviation of the error as 2- and 1-norms.
We also employed $R^2$ as RMSE and MAE do not account for the trend between real and predicted values.

We used the Electricity Transformer Temperature (ETT) dataset collected from two transformers in two Chinese provinces, one each \cite{Informer}.
Indicators m or h represent intervals of data collected as 15 min or 1 h, respectively.
For example, the 15-min interval dataset collected from the first transformer is ETTm1, and each from ETT contains seven time series. The senven time series contain the ``UseFul" and ``UseLess" Load with three types of power loads (i.e. high, middle and low), and oil temperature. For convenience, we will use abbreviations: high useful loads, high useless loads, middle useful loads, middle useless loads, low useful loads, and low useless loads are HUFL, HULL, MUFL, MULL, LUFL, and LULL, respectively. The ETTh1 and ETTh2 has 17,421 time lengths while ETTm1 and ETTm2 has 69,681 time lengths. 

% The details of the dataset are listed in Table \ref{tab:date_config}.

% \usepackage{multirow}
% \usepackage{graphicx}
% \begin{table}[!t]
% \caption{The details of the datasets. BAQ has the same shape for all time series datasets.}
% \label{tab:date_config}
% \vspace{10pt}
% \centering
% \resizebox{.7\textwidth}{!}{%
% \begin{tabular}{lcllcl}\toprule
% \multicolumn{3}{c}{ETT}               & \multicolumn{3}{c}{M4}                 \\ \midrule
% Dataset & \# of Time series  & Length & Dataset   & \# of Time series & Length \\ \midrule
% ETTh1   & \multirow{4}{*}{7} & 17421  & Daily     & 93                & 4228   \\
% ETTh2   &                    & 17421  & Monthly   & 43                & 48001  \\
% ETTm1   &                    & 69681  & Quarterly & 16                & 24001  \\
% ETTm2   &                    & 69681  & Yearly    & 13                & 23001  \\ \bottomrule
% \end{tabular}%
% }
% \end{table}

We compared NAM-NCs' performance to three comparative models. 
Table \ref{tab:ex_ett} shows the experimental NC results for the ETT dataset. 
The first row indicates the dataset and metric types. 
The rows below show the metrics of the method used by the province transformer per time interval. 
% renewal paragraph
In the metric tables, the \textbf{bold} and \underline{underline} indicates the best scores of all NN methods and NAM-NCs' variations, respectively.
NAM-NCs showed the best performance from the second transformer.
NAM-NCs with shared parameters resulted in lower scores in terms of $R^2$ compared to the original NAM-NC.
However, NAM-NC$_{feature}$ showed the best performacnes in ETTh2 and ETTm2 in terms of RMSE and MAE.
Also, NAM-NC$_{time}$ showed the best $R^2$ in ETTm2.
In particular, all NAM-NCs often outperformed other comparison methods.
SCINet maintained the performance consistency.
SCINet achieved the best performance for all metrics from the first transformer. 
Meanwhile, LSTM showed the worst performance.

% editing paragraph
% NAM-NC showed the best performance from the second transformer.
% In addition to $R^2$, other metrics well-represented NC.
% Lightweight NAM-NC$_{feature}$ performance decreased the $R^2$ by 0.01, compared with NAM-NC, but it showed better results in absolute deviation metrics. 
% Moreover, for ETTm2, the NAM-NC with shared features showed the best metrics. 
% SCINet showed the best performance in both ETTh1 and ETTm1 datasets, and NAM-NC ranked second. 
% The Informer paper originally presented the ETT dataset, and our NAM-NC method showed similar results. 
% LSTM showed the worst performance among all tested methods.

% \usepackage{multirow}
% \usepackage{graphicx}
\begin{table}[!t]
\centering
\caption{Summary of the experiment for ETT dataset}
\label{tab:ex_ett}
\vspace{10pt}
\resizebox{\textwidth}{!}{%
\begin{tabular}{lllrrrrllrrrr} \toprule
Category              & Dataset                & Method             & $\#$ Params & $R^2$     & RMSE   & MAE    & Dataset                & Method             & $\#$ Params & $R^2$     & RMSE   & MAE    \\ \midrule
\multirow{12}{*}{ETT} & \multirow{6}{*}{ETTh1} & NAM-NC             & 193k          & \underline{0.7474} & \underline{1.4004} & \underline{0.7954} & \multirow{6}{*}{ETTm1} & NAM-NC             & 193k          & \underline{0.8997} & \underline{0.8350} & \underline{0.4424} \\
                      &                        & NAM-NC$_{time}$    & 24k           & 0.7434 & 1.4172 & 0.8065 &                        & NAM-NC$_{time}$    & 24k           & 0.8932 & 0.8571 & 0.4553 \\
                      &                        & NAM-NC$_{feature}$ & 28k           & 0.7432 & 1.4054 & 0.7973 &                        & NAM-NC$_{feature}$ & 28k           & 0.8986 & 0.8442 & 0.4466 \\
                      &                        & Informer           & 330k          & 0.7407 & 1.3598 & 0.7860 &                        & Informer           & 330k          & 0.8874 & 0.8946 & 0.4736 \\
                      &                        & LSTM               & 188k          & 0.7006 & 1.4385 & 0.8387 &                        & LSTM               & 188k          & 0.8793 & 0.8950 & 0.4856 \\
                      &                        & SCINet             & 11k           & \textbf{0.7695} & \textbf{1.3093} & \textbf{0.7338} &                        & SCINet             & 11k           & \textbf{0.9037} & \textbf{0.8092} & \textbf{0.4168} \\ \cline{2-13}
                      & \multirow{6}{*}{ETTh2} & NAM-NC             & 193k          & \textbf{\underline{0.6289}} & 2.0359 & 1.2897 & \multirow{6}{*}{ETTm2} & NAM-NC             & 193k          & 0.8250 & 1.3256 & 0.7981 \\
                      &                        & NAM-NC$_{time}$    & 24k           & 0.6162 & 2.0188 & 1.3013 &                        & NAM-NC$_{time}$    & 24k           & \textbf{\underline{0.8272}} & 1.2147 & 0.7750 \\
                      &                        & NAM-NC$_{feature}$ & 28k           & 0.6134 & \textbf{\underline{1.9767}} & \textbf{\underline{1.2847}} &                        & NAM-NC$_{feature}$ & 27k           & 0.8259 & \textbf{\underline{1.1950}} & \textbf{\underline{0.7591}} \\
                      &                        & Informer           & 330k          & 0.5856 & 2.3137 & 1.4151 &                        & Informer           & 330k          & 0.7781 & 1.6681 & 0.9368 \\
                      &                        & LSTM               & 188k          & 0.5041 & 2.7676 & 1.7218 &                        & LSTM               & 188k          & 0.7582 & 1.7746 & 1.0120 \\
                      &                        & SCINet             & 11k           & 0.6123 & 2.1666 & 1.3356 &                        & SCINet             & 11k           & 0.8050 & 1.4351 & 0.8272 \\ \bottomrule
\end{tabular}%
}
\end{table}

\begin{figure*}[!th]
    \centering
    \begin{subfigure}[b]{0.48\textwidth}
        \centering
        \includegraphics[width=\textwidth]{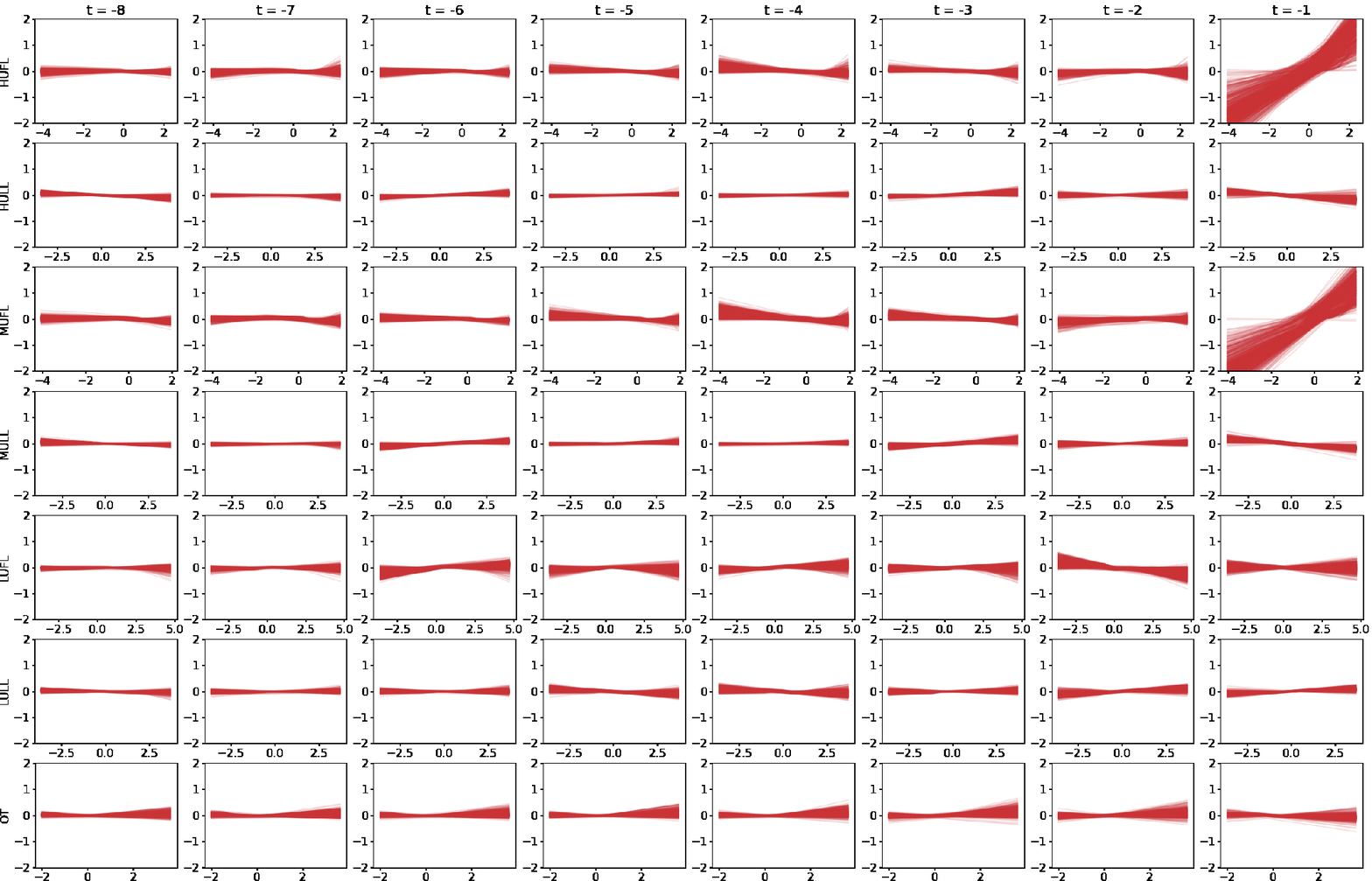}
        \caption{HUFL}
    \end{subfigure}
    \hfill
    \begin{subfigure}[b]{0.48\textwidth}
        \centering
        \includegraphics[width=\textwidth]{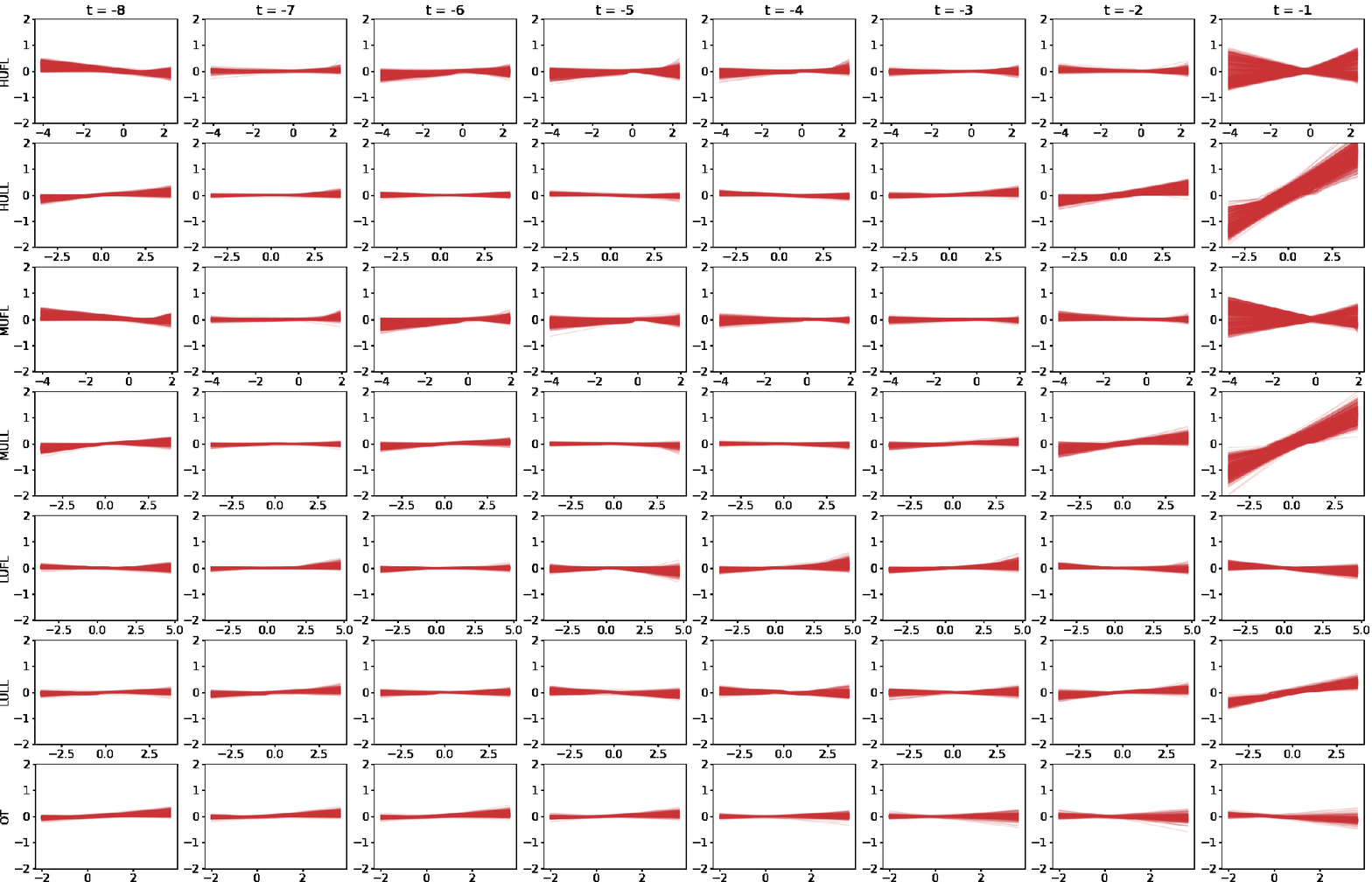}
        \caption{HULL}
    \end{subfigure}
    \vspace{5pt}
    \begin{subfigure}[b]{0.48\textwidth}
        \centering
        \includegraphics[width=\textwidth]{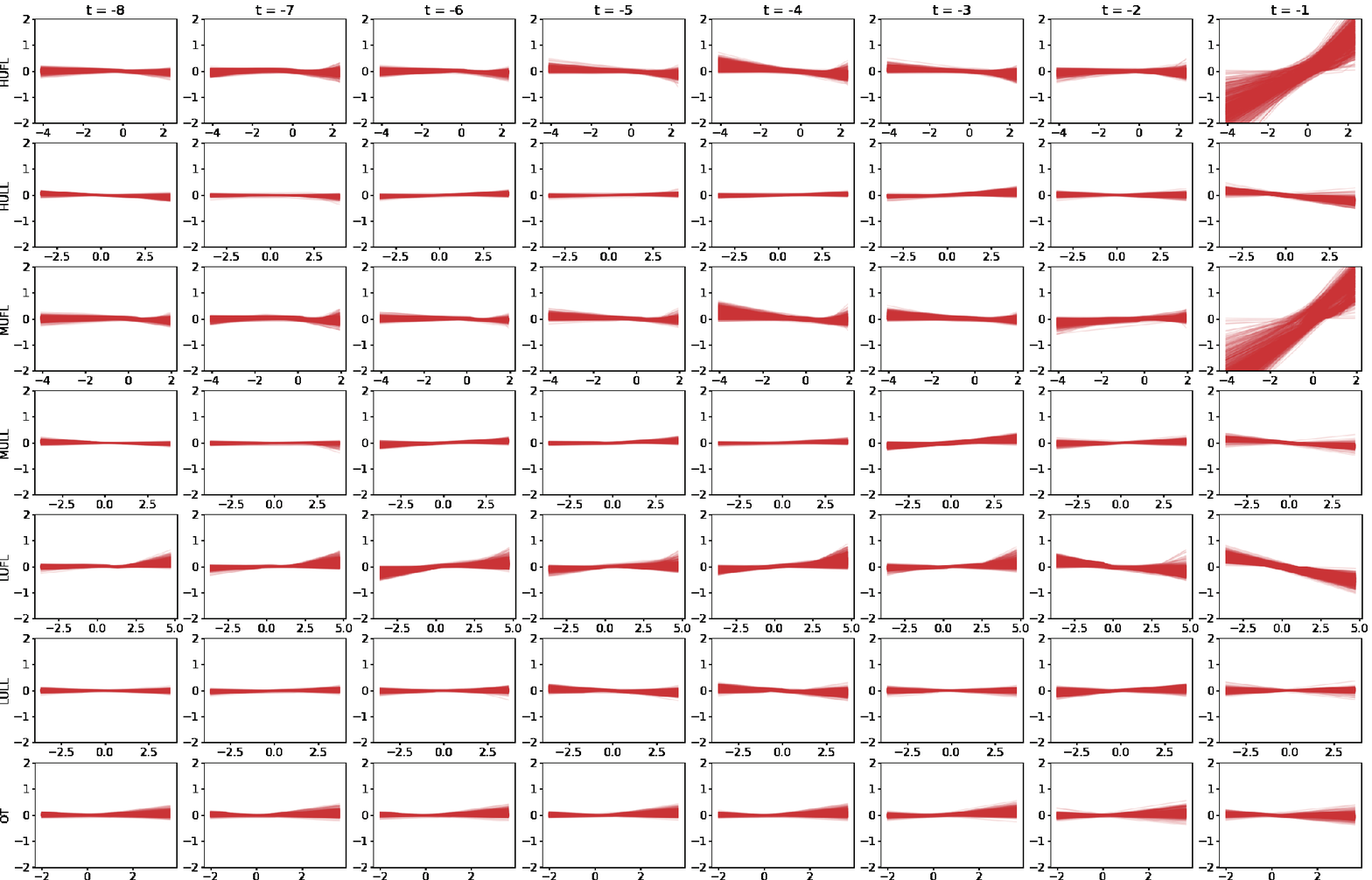}
        \caption{MUFL}
    \end{subfigure}
    \hfill
    \begin{subfigure}[b]{.48\textwidth}
        \centering
        \includegraphics[width=\textwidth]{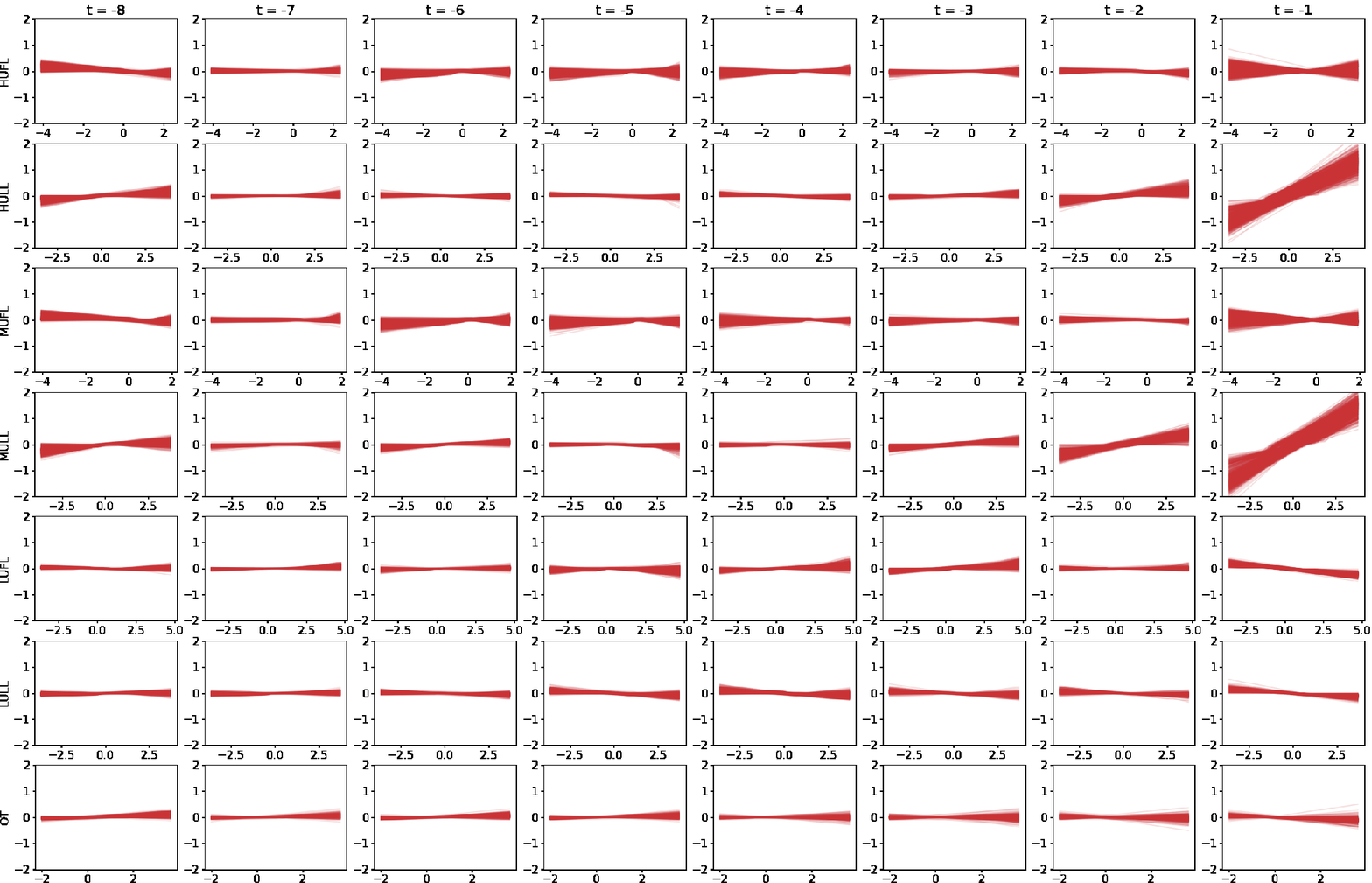}
        \caption{MULL}
    \end{subfigure}
    \vspace{5pt}
    \begin{subfigure}[b]{.48\textwidth}
        \centering
        \includegraphics[width=\textwidth]{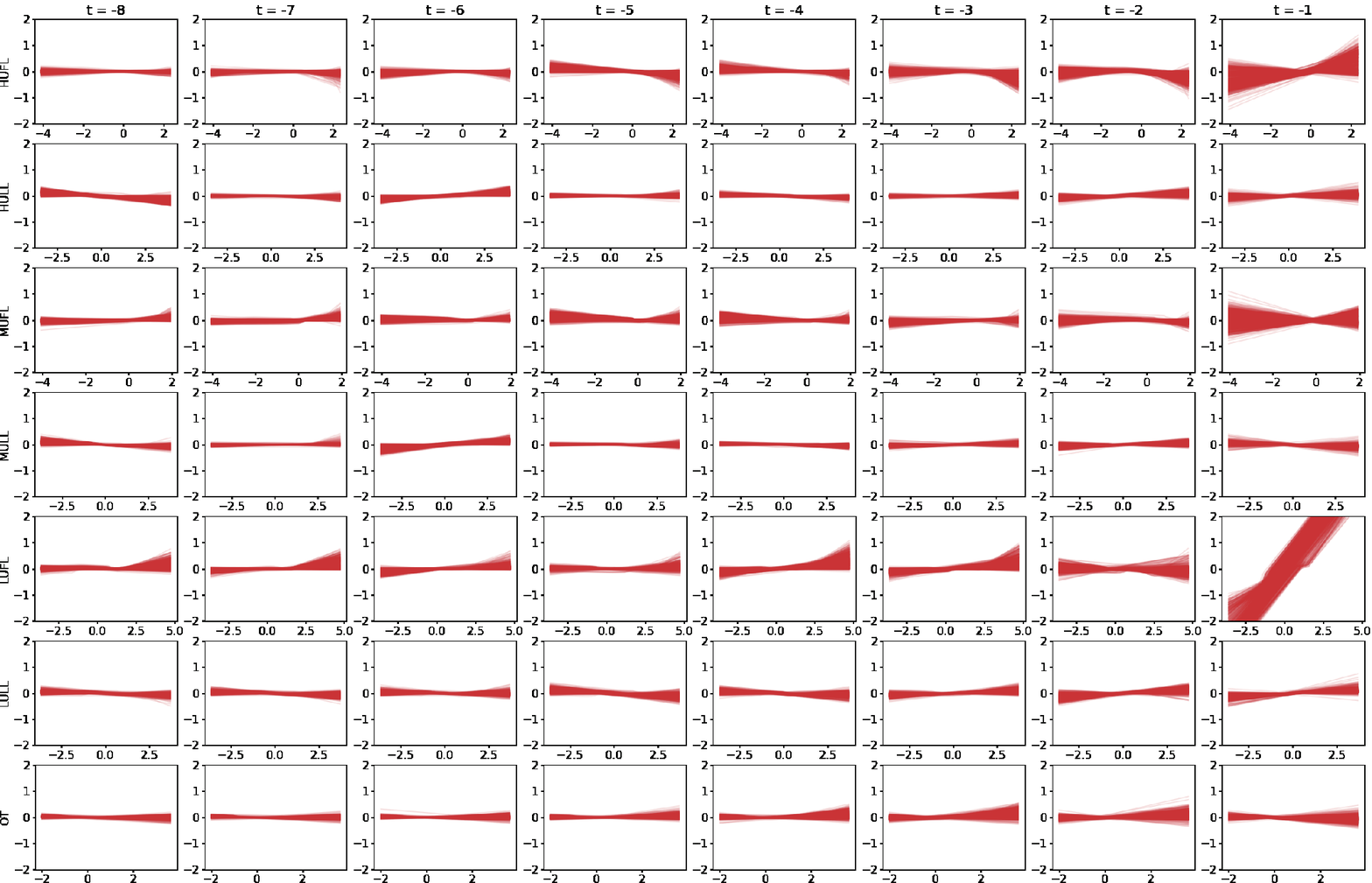}
        \caption{LUFL}
    \end{subfigure}
    \hfill
    \begin{subfigure}[b]{.48\textwidth}
        \centering
        \includegraphics[width=\textwidth]{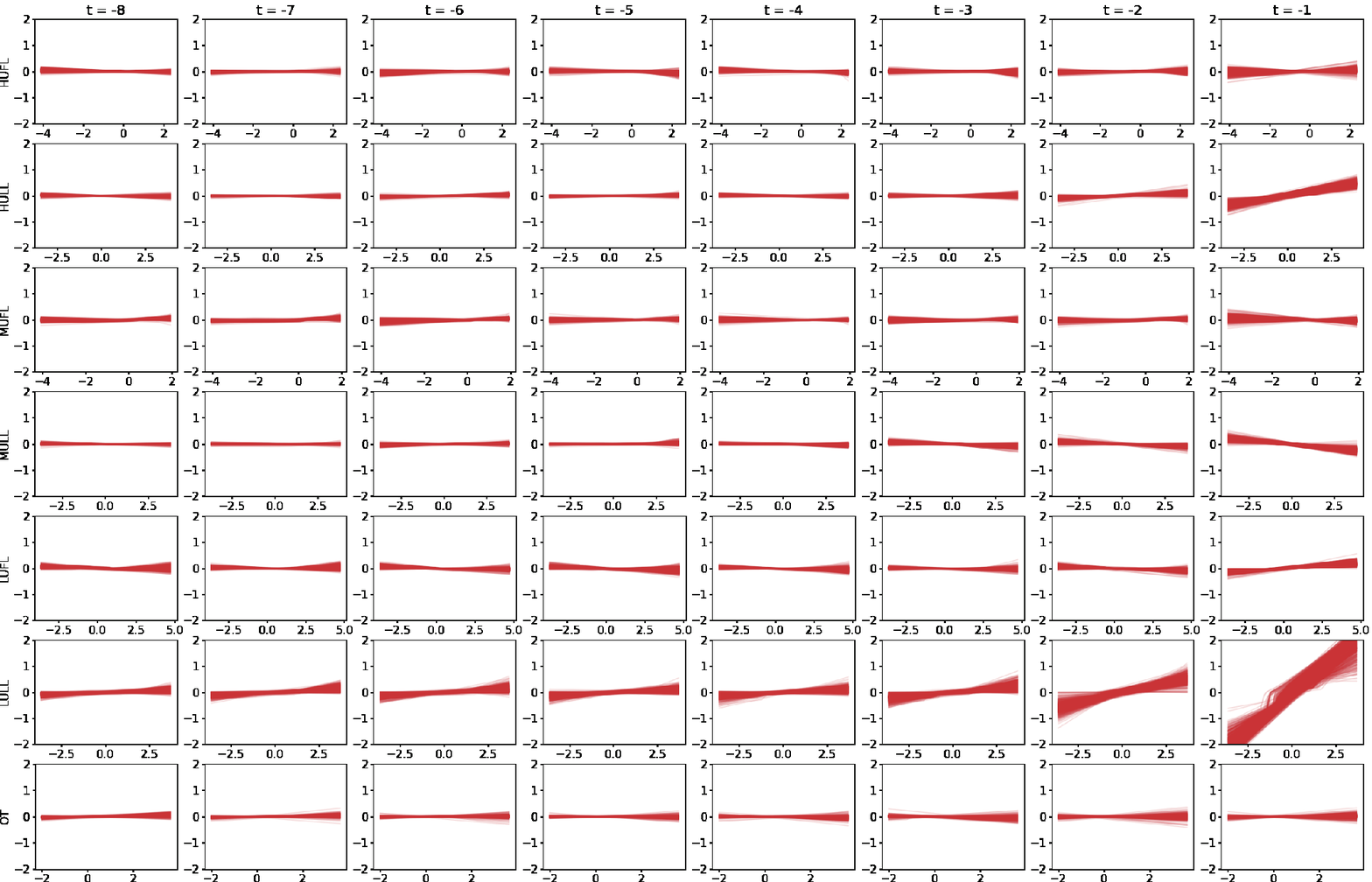}
        \caption{LULL}
    \end{subfigure}
     
    \caption{Contribution of the unique values in time series of the ETTh1 datasets when to predict the next time step's value. The subfigure's caption indicates the nowcasting target.}
    \label{fig:ett_cont}
\end{figure*}

\begin{figure*}[!t]
\centering
\includegraphics[width = .9\textwidth]{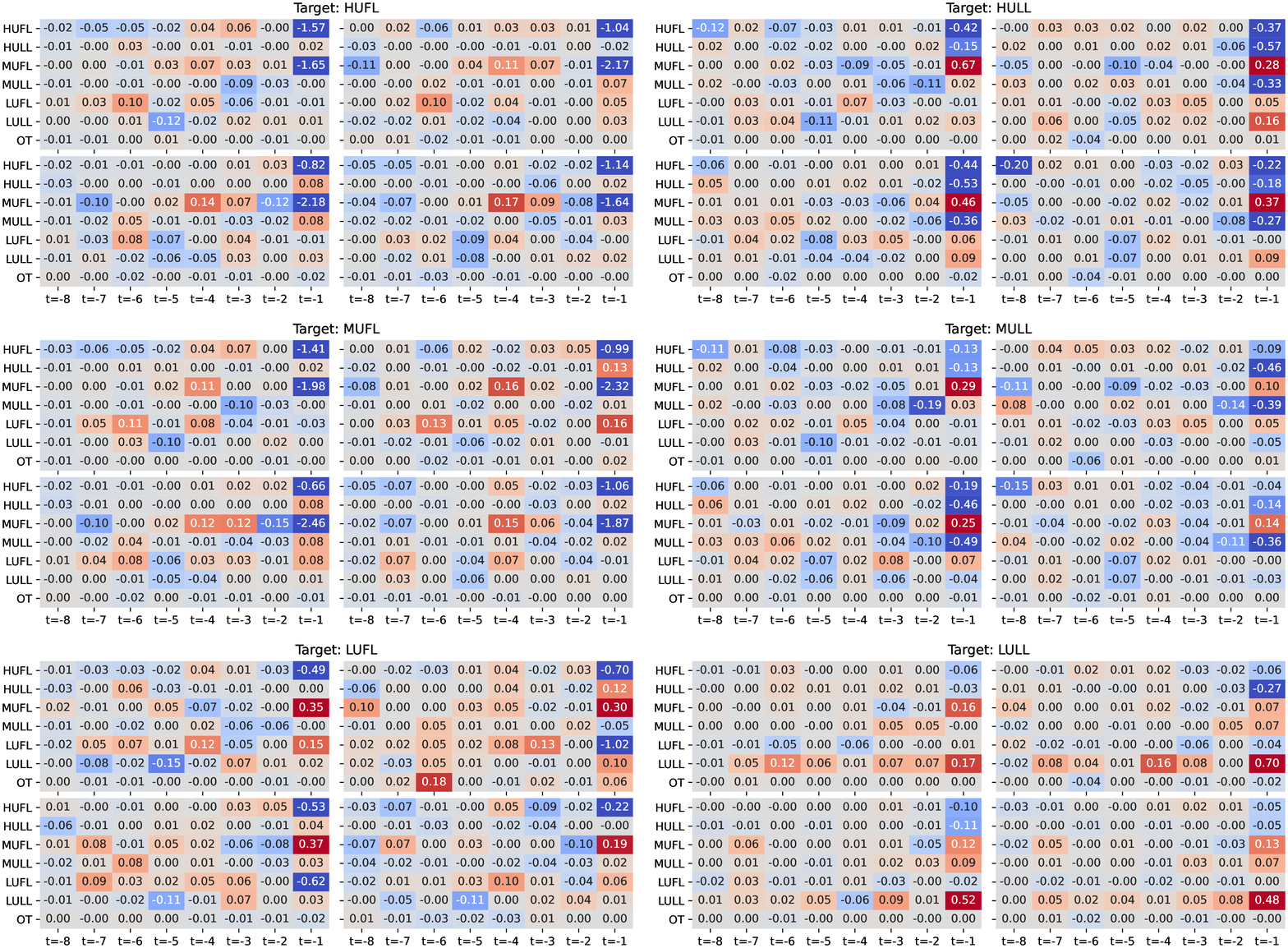}
\caption{Importance value of each input variable for each target of ETTh1.}
\label{fig:cont_ett_sample}
\end{figure*}

% \usepackage{multirow}
% \usepackage{graphicx}
\begin{table}[!t]
\centering
\caption{Summary of the experiment for BAQ time series datasets.}
\label{tab:other_exp_result}
\vspace{10pt}
\resizebox{\textwidth}{!}{%
\begin{tabular}{lllrrrrllrrrr}\toprule
Category              & Dataset                       & Method             & $\#$ Params & $R^2$      & RMSE    & MAE     & Dataset                        & Method             & $\#$ Params & $R^2$     & RMSE    & MAE     \\\midrule
\multirow{36}{*}{BAQ} & \multirow{6}{*}{Aotizhongxin} & NAM-NC             & 275k   & 0.8651  & 105.23  & 23.21   & \multirow{6}{*}{Nongzhanhuan}  & NAM-NC             & 275k   & 0.8742 & 105.73  & 23.19   \\
                      &                               & NAM-NC$_{time}$    & 35k    & 0.8599  & 105.27  & 23.35   &                                & NAM-NC$_{time}$    & 35k    & 0.8719 & 106.34  & 23.61   \\
                      &                               & NAM-NC$_{feature}$ & 28k    & \underline{0.8669}  & \underline{104.88}  & \underline{23.10}   &                                & NAM-NC$_{feature}$ & 28k    & \underline{0.8760} & \underline{105.63}  & \underline{23.02}   \\
                      &                               & Informer           & 333k   & 0.8569  & 112.36  & 24.21   &                                & Informer           & 333k   & 0.8705 & 110.60  & 24.49   \\
                      &                               & LSTM               & 189k   & 0.8538  & 107.86  & 23.83   &                                & LSTM               & 189k   & 0.8696 & 106.98  & 23.72   \\
                      &                               & SCINet             & 23k    & \textbf{0.8768}  & \textbf{103.64}  & \textbf{22.04}   &                                & SCINet             & 23k    & \textbf{0.8864} & \textbf{103.62}  & \textbf{21.83}   \\\cline{2-13}
                      & \multirow{6}{*}{Changping}    & NAM-NC             & 275k   & 0.8337  & 110.00  & 24.95   & \multirow{6}{*}{Shunyi}        & NAM-NC             & 275k   & 0.8438 & 108.42  & 23.53   \\
                      &                               & NAM-NC$_{time}$    & 35k    & 0.8321  & 110.38  & 25.14   &                                & NAM-NC$_{time}$    & 35k    & 0.8514 & \textbf{\underline{106.38}}  & 23.08   \\
                      &                               & NAM-NC$_{feature}$ & 28k    & \underline{0.8372}  & \underline{109.75}  & \underline{24.68}   &                                & NAM-NC$_{feature}$ & 28k    & \textbf{\underline{0.8547}} & 107.17  & \underline{22.91}   \\
                      &                               & Informer           & 333k   & 0.8356  & 109.56  & 25.09   &                                & Informer           & 333k   & 0.8337 & 116.99  & 24.59   \\
                      &                               & LSTM               & 189k   & 0.8323  & 109.61  & 24.66   &                                & LSTM               & 189k   & 0.8184 & 115.98  & 24.43   \\
                      &                               & SCINet             & 23k    & \textbf{0.8493}  & \textbf{106.10}  & \textbf{23.36}   &                                & SCINet             & 23k    & 0.8532 & 108.34  & \textbf{22.21}   \\\cline{2-13}
                      & \multirow{6}{*}{Dingling}     & NAM-NC             & 275k   & 0.8633  & 76.98   & 16.99   & \multirow{6}{*}{Tiantan}       & NAM-NC             & 275k   & \underline{0.8544} & 106.69  & 23.06   \\
                      &                               & NAM-NC$_{time}$    & 35k    & 0.8607  & 77.15   & 17.34   &                                & NAM-NC$_{time}$    & 35k    & 0.8501 & 106.68  & 23.57   \\
                      &                               & NAM-NC$_{feature}$ & 28k    & \underline{0.8644}  & \underline{76.86}   & \underline{16.84}   &                                & NAM-NC$_{feature}$ & 28k    & 0.8540 & \underline{106.15}  & \underline{22.95}   \\
                      &                               & Informer           & 333k   & 0.8588  & 78.78   & 17.55   &                                & Informer           & 333k   & 0.8512 & 106.63  & 23.11   \\
                      &                               & LSTM               & 189k   & 0.8565  & 80.31   & 18.13   &                                & LSTM               & 189k   & 0.8507 & 106.22  & 23.17   \\
                      &                               & SCINet             & 23k    & \textbf{0.8758}  & \textbf{75.64}   & \textbf{15.86}   &                                & SCINet             & 23k    & \textbf{0.8667} & \textbf{102.64}  & \textbf{21.64}   \\\cline{2-13}
                      & \multirow{6}{*}{Dongsi}       & NAM-NC             & 275k   & -0.4103 & 94.25   & 21.30   & \multirow{6}{*}{Wanliu}        & NAM-NC             & 275k   & 0.8611 & 97.05   & \underline{21.80}   \\
                      &                               & NAM-NC$_{time}$    & 35k    & -0.5088 & 94.85   & 21.61   &                                & NAM-NC$_{time}$    & 35k    & 0.8582 & 99.02   & 22.86   \\
                      &                               & NAM-NC$_{feature}$ & 28k    & \underline{0.0755}  & \underline{93.72}   & \underline{20.88}   &                                & NAM-NC$_{feature}$ & 28k    & \underline{0.8650} & \underline{97.02}   & 22.09   \\
                      &                               & Informer           & 333k   & -1.3288 & 97.13   & 22.02   &                                & Informer           & 333k   & 0.8516 & 100.70  & 23.44   \\
                      &                               & LSTM               & 189k   & 0.1420  & 99.00   & 22.40   &                                & LSTM               & 189k   & 0.8496 & 103.30  & 24.19   \\
                      &                               & SCINet             & 23k    & \textbf{0.4657}  & \textbf{92.85}   & \textbf{19.99}   &                                & SCINet             & 23k    & \textbf{0.8755} & \textbf{93.91}   & \textbf{20.61}   \\\cline{2-13}
                      & \multirow{6}{*}{Guanyuan}     & NAM-NC             & 275k   & 0.8670  & 95.32   & 21.05   & \multirow{6}{*}{Wanshouxigong} & NAM-NC             & 275k   & 0.8525 & \underline{104.28}  & 24.76   \\
                      &                               & NAM-NC$_{time}$    & 35k    & 0.8665  & \underline{94.54}   & 21.25   &                                & NAM-NC$_{time}$    & 35k    & 0.8500 & 104.57  & 24.92   \\
                      &                               & NAM-NC$_{feature}$ & 28k    & \underline{0.8702}  & 94.90   & \underline{21.03}   &                                & NAM-NC$_{feature}$ & 28k    & \underline{0.8535} & 104.54  & \underline{24.63}   \\
                      &                               & Informer           & 333k   & 0.8658  & 99.45   & 22.02   &                                & Informer           & 333k   & 0.8472 & 104.83  & 24.95   \\
                      &                               & LSTM               & 189k   & 0.8621  & 96.93   & 21.81   &                                & LSTM               & 189k   & 0.8433 & 103.80  & 24.78   \\
                      &                               & SCINet             & 23k    & \textbf{0.8798}  & \textbf{92.15}   & \textbf{19.70}   &                                & SCINet             & 23k    & \textbf{0.8661} & \textbf{100.32}  & \textbf{23.02}   \\\cline{2-13}
                      & \multirow{6}{*}{Gucheng}      & NAM-NC             & 275k   & 0.8353  & 99.85   & 24.16   & \multirow{6}{*}{Huairou}       & NAM-NC             & 275k   & \underline{0.8415} & \underline{82.04}   & \underline{19.21}   \\
                      &                               & NAM-NC$_{time}$    & 35k    & 0.8362  & 99.78   & 24.26   &                                & NAM-NC$_{time}$    & 35k    & 0.8369 & 82.23   & 19.63   \\
                      &                               & NAM-NC$_{feature}$ & 28k    & \underline{0.8440}  & \underline{99.50}   & \underline{23.83}   &                                & NAM-NC$_{feature}$ & 28k    & 0.8400 & 82.23   & 19.23   \\
                      &                               & Informer           & 333k   & 0.8367  & 100.95  & 25.03   &                                & Informer           & 333k   & 0.8350 & 84.02   & 19.43   \\
                      &                               & LSTM               & 189k   & 0.8315  & 104.70  & 25.79   &                                & LSTM               & 189k   & 0.8353 & 82.22   & 19.52   \\
                      &                               & SCINet             & 23k    & \textbf{0.8530}  & \textbf{97.76}   & \textbf{22.48}   &                                & SCINet             & 23k    & \textbf{0.8567} & \textbf{79.68}   & \textbf{18.04}  \\ \bottomrule
\end{tabular}%
}
\end{table}

Unlike other methods, our model can explain how an NN makes its predictions. 
Figure \ref{fig:ett_cont} visualizes the scalar extracted by NAM-NC’s feature net from the ETTh1 dataset. 
The feature scalars of the unique values of each time series in ETTh1 extracted by the feature net revealed large deviations from the input time series immediately prior to NC. 
We also found a strong correlation between HUFL and MUFL and between HULL and MULL, according to the unique inputs. 
We further confirmed a similar pattern in the time-series samples of ETTh1. 
Figure \ref{fig:cont_ett_sample} shows the contributions generated by the feature net of NAM-NC trained four times to repeat each NC target. 
According to the figure, each trained model extracted contributions by ignoring unimportant time-series points and focusing on critical points. 
Furthermore, NAM-NC did not always obtain the contribution information from the immediately preceding item; it also extracted contributions from the distant past.

\begin{figure*}[!t]
\centering
\includegraphics[width = .9\textwidth]{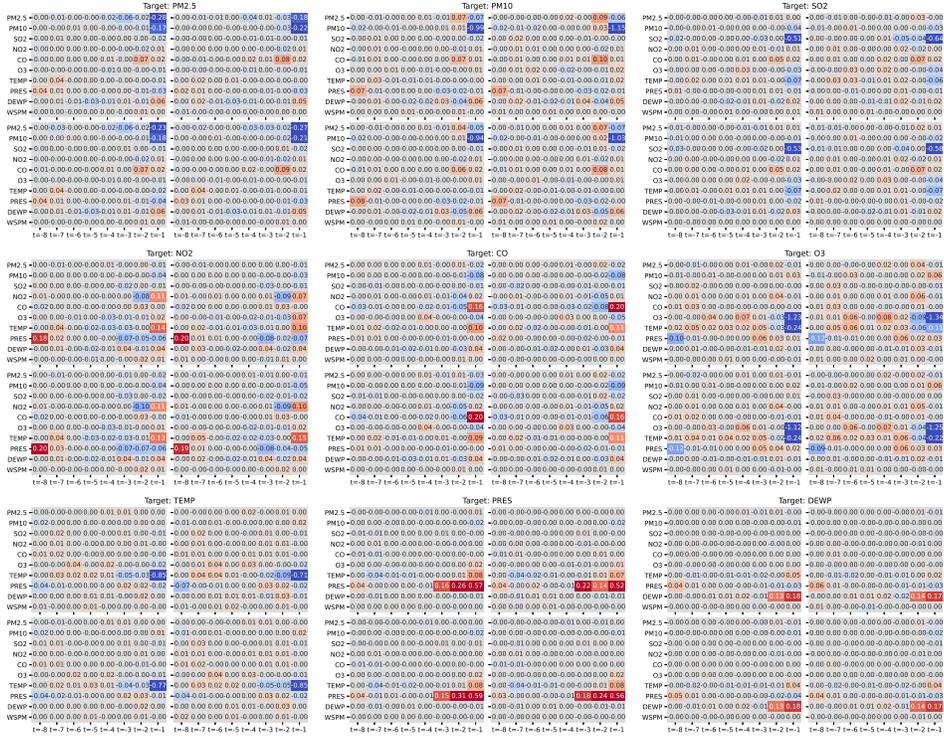}
\caption{Importance value of each input variable for each target in BAQ's Shunyi site.}
\label{fig:cont_baq_sample}
\end{figure*}

We employed another dataset: Beijing Air Quality (BAQ) \cite{baq_dataset}. 
All BAQ time-series data consist of 10 time series contain to air quality, $SO_2$, and more. 
Objects with time lengths of 35,065, the size of fine dust, are contained in the set, including temperature, humidity, etc., taken from monitoring sites near Beiging.
% However, for M4, we removed time series with missing values. 
% The dataset details for M4 are found in Table \ref{tab:date_config}.
According to Table \ref{tab:other_exp_result}, NAM-NC$_{feature}$ performed the best in most metrics among NAM-NCs. It is seemed that NAM-NCs with shared parameters provide computational efficiency, but do not degraded the prediction performance. SCINet showed the best prediction performance for most datasets; however the NAM-NCs showed comparable results while providing the explanation power.
Figure \ref{fig:cont_baq_sample} shows the contributions generated by the feature net of NAM-NC trained four times to repeat each NC target of BAQ.
According to the figure, NAM-NC captured the time series which is correlated, as temperature, pressure, and dew point. Also, we explained the NAM-NC's prediction for the $NO_2$ and $O_3$ that received the contribution from the distant past, such as pressure at $t=-8$.

% According to Table \ref{tab:other_exp_result}, which summarizes our experimental results, LSTM performed the best with M4, and NAM-NC showed the second-best performance. 
% Owing to M4’s NC process, the absolute deviation of error was large in all models. 
% With BAQ, SCINet performed the best, and NAM-NC$_{feature}$ ranked second in most metrics. 
% However, it performed poorly on the Dongsi site data, where both Informer and NAM-NC produced $R^2$ values less than zero.
% SCINet outperformed the other models, followed by LSTM. 

\section{Conclusion}\label{sec.6}

Despite its nonrecursive structure, NAM-NC supports NC tasks well. 
Owing to its GAM basis, NAM-NC can explain prediction results on both synthetic and real datasets. 
We used hard sharing to reduce the number of parameters, and NAM-NC performed just as well as before parameter grouping, although the number of parameters was considerably reduced. 
To show the explainability of NAM-NC, we visualized its predictions using feature nets for unique values in synthetic data. 
Both synthetic and real time-series data were used, and NAM-NC extracted the tendency contributions of the unique time-series inputs. 
Even with repeated learning, we confirmed the model’s consistency in extracting most of the large contributions from the important points identified in the time series data. 
Therefore, NAM-NC has a competitive edge over other NN models due to its explainability under the assumption that it learns and predicts fields that support NC tasks.

NAM-NC still faces several obstacles. 
In this study, hyperparameters were not specifically explored, and we did not utilize the various time-series lengths. 
Therefore, it is necessary to verify NAM-NC with other types of datasets. 
The model showed good performance in the ETT and BAQ problems, which have strong periodicity. 
% However, it showed poor performance with M4, which has weak periodicity. 
Thus, the recursion of NNs based on scalars extracted by feature nets requires more attention. 
Additionally, researchers should expand the scope of NAM-NC beyond recursion into forecasting.

\bibliography{bib}

\begin{thebibliography}{10}

\bibitem{electra}
Clark, K., M.-T. Luong, Q.~V. Le, et~al.
\newblock {ELECTRA}: Pre-training text encoders as discriminators rather than
  generators.
\newblock In \emph{ICLR}. 2020.

\bibitem{yolov3}
Redmon, J., A.~Farhadi.
\newblock Yolov3: An incremental improvement.
\newblock \emph{ArXiv}, abs/1804.02767, 2018.

\bibitem{TCN}
Lea, C., M.~D. Flynn, R.~Vidal, et~al.
\newblock Temporal convolutional networks for action segmentation and
  detection.
\newblock In \emph{2017 IEEE Conference on Computer Vision and Pattern
  Recognition (CVPR)}, pages 1003--1012. 2017.

\bibitem{adv_nn}
Tu, J.~V.
\newblock Advantages and disadvantages of using artificial neural networks
  versus logistic regression for predicting medical outcomes.
\newblock \emph{J Clin Epidemiol}, 49(11):1225--1231, 1996.

\bibitem{Informer}
Zhou, H., S.~Zhang, J.~Peng, et~al.
\newblock Informer: Beyond efficient transformer for long sequence time-series
  forecasting.
\newblock In \emph{AAAI}. 2021.

\bibitem{rainfall}
Ridwan, W.~M., M.~Sapitang, A.~Aziz, et~al.
\newblock Rainfall forecasting model using machine learning methods: Case study
  terengganu, malaysia.
\newblock \emph{Ain Shams Engineering Journal}, 12(2):1651--1663, 2021.

\bibitem{nam}
Agarwal, R., L.~Melnick, N.~Frosst, et~al.
\newblock Neural additive models: Interpretable machine learning with neural
  nets.
\newblock In A.~Beygelzimer, Y.~Dauphin, P.~Liang, J.~W. Vaughan, eds.,
  \emph{Advances in Neural Information Processing Systems}. 2021.

\bibitem{gam}
Hastie, T., R.~Tibshirani.
\newblock Generalized additive models.
\newblock \emph{Statistical Science}, 1(3):297--310, 1986.
\newblock Full publication date: Aug., 1986.

\bibitem{ar_theory}
Walker, G.~T.
\newblock On periodicity in series of related terms.
\newblock \emph{Proceedings of The Royal Society A: Mathematical, Physical and
  Engineering Sciences}, 131:518--532, 1931.

\bibitem{vectorAR}
Stock, J.~H., M.~W. Watson.
\newblock Vector autoregressions.
\newblock \emph{Journal of Economic Perspectives}, 15(4):101--115, 2001.

\bibitem{ar_nowcasting}
Carriero, A., T.~E. Clark, M.~Marcellino.
\newblock Realtime nowcasting with a bayesian mixed frequency model with
  stochastic volatility.
\newblock \emph{Journal of the Royal Statistical Society. Series A, (Statistics
  in Society)}, 178(4):837--862, 2015.
\newblock 27840562[pmid].

\bibitem{prophet}
Taylor, S.~J., B.~Letham.
\newblock Forecasting at scale.
\newblock \emph{The American Statistician}, 72(1):37--45, 2018.

\bibitem{ARnet}
Zhu, Y., C.~Chen, G.~Yan, et~al.
\newblock Ar-net: Adaptive attention and residual refinement network for
  copy-move forgery detection.
\newblock \emph{IEEE Transactions on Industrial Informatics},
  16(10):6714--6723, 2020.

\bibitem{arconv}
Binkowski, M., G.~Marti, P.~Donnat.
\newblock Autoregressive convolutional neural networks for asynchronous time
  series.
\newblock In J.~Dy, A.~Krause, eds., \emph{Proceedings of the 35th
  International Conference on Machine Learning}, vol.~80, pages 580--589. PMLR,
  2018.

\bibitem{neural_prophet}
Triebe, O., H.~Hewamalage, P.~Pilyugina, et~al.
\newblock Neuralprophet: Explainable forecasting at scale, 2021.

\bibitem{gbmnowcast1}
Soybilgen, B., E.~Yazgan.
\newblock Nowcasting us gdp using tree-based ensemble models and dynamic
  factors.
\newblock \emph{Computational Economics}, 57(1):387--417, 2021.

\bibitem{lstm}
Hochreiter, S., J.~Schmidhuber.
\newblock Long short-term memory.
\newblock \emph{Neural Comput.}, 9(8):1735–1780, 1997.

\bibitem{lstma}
Bahdanau, D., K.~Cho, Y.~Bengio.
\newblock Neural machine translation by jointly learning to align and
  translate.
\newblock \emph{arXiv preprint arXiv:1409.0473}, 2014.

\bibitem{hst-lstm}
Kong, D., F.~Wu.
\newblock Hst-lstm: A hierarchical spatial-temporal long-short term memory
  network for location prediction.
\newblock In \emph{Proceedings of the 27th International Joint Conference on
  Artificial Intelligence}, page 2341–2347. AAAI Press, 2018.

\bibitem{lstnet}
Lai, G., W.-C. Chang, Y.~Yang, et~al.
\newblock Modeling long- and short-term temporal patterns with deep neural
  networks.
\newblock \emph{The 41st International ACM SIGIR Conference on Research \&
  Development in Information Retrieval}, 2018.

\bibitem{attention}
Vaswani, A., N.~Shazeer, N.~Parmar, et~al.
\newblock Attention is all you need.
\newblock In I.~Guyon, U.~V. Luxburg, S.~Bengio, H.~Wallach, R.~Fergus,
  S.~Vishwanathan, R.~Garnett, eds., \emph{Advances in Neural Information
  Processing Systems}, vol.~30. Curran Associates, Inc., 2017.

\bibitem{cnn}
Krizhevsky, A., I.~Sutskever, G.~E. Hinton.
\newblock Imagenet classification with deep convolutional neural networks.
\newblock In F.~Pereira, C.~Burges, L.~Bottou, K.~Weinberger, eds.,
  \emph{Advances in Neural Information Processing Systems}, vol.~25. Curran
  Associates, Inc., 2012.

\bibitem{bert}
Devlin, J., M.-W. Chang, K.~Lee, et~al.
\newblock {BERT}: Pre-training of deep bidirectional transformers for language
  understanding.
\newblock In \emph{Proceedings of the 2019 Conference of the North {A}merican
  Chapter of the Association for Computational Linguistics: Human Language
  Technologies, Volume 1 (Long and Short Papers)}, pages 4171--4186.
  Association for Computational Linguistics, Minneapolis, Minnesota, 2019.

\bibitem{positionale}
Ke, G., D.~He, T.-Y. Liu.
\newblock Rethinking positional encoding in language pre-training.
\newblock In \emph{International Conference on Learning Representations}. 2021.

\bibitem{dilated}
Yu, F., V.~Koltun.
\newblock Multi-scale context aggregation by dilated convolutions.
\newblock In \emph{International Conference on Learning Representations
  (ICLR)}. 2016.

\bibitem{scinet}
Liu, M., A.~Zeng, Q.~Lai, et~al.
\newblock Time series is a special sequence: Forecasting with sample
  convolution and interaction.
\newblock \emph{ArXiv}, abs/2106.09305, 2021.

\bibitem{exai2}
Montavon, G., W.~Samek, K.-R. Muller.
\newblock Methods for interpreting and understanding deep neural networks.
\newblock \emph{Digital Signal Processing}, 73:1--15, 2018.

\bibitem{exai1}
Linardatos, P., V.~Papastefanopoulos, S.~Kotsiantis.
\newblock Explainable ai: A review of machine learning interpretability
  methods.
\newblock \emph{Entropy}, 23(1), 2021.

\bibitem{cam}
Zhou, B., A.~Khosla, L.~A., et~al.
\newblock {Learning Deep Features for Discriminative Localization.}
\newblock \emph{CVPR}, 2016.

\bibitem{exrnn}
Shih, S.-Y., F.-K. Sun, H.-y. Lee.
\newblock Temporal pattern attention for multivariate time series forecasting.
\newblock \emph{Machine Learning}, 108(8):1421--1441, 2019.

\bibitem{lrp}
Lapuschkin, S., A.~Binder, G.~Montavon, et~al.
\newblock On pixel-wise explanations for non-linear classifier decisions by
  layer-wise relevance propagation.
\newblock \emph{PLoS ONE}, 10:e0130140, 2015.

\bibitem{lime}
Ribeiro, M., S.~Singh, C.~Guestrin.
\newblock {``}why should {I} trust you?{''}: Explaining the predictions of any
  classifier.
\newblock In \emph{Proceedings of the 2016 Conference of the North {A}merican
  Chapter of the Association for Computational Linguistics: Demonstrations},
  pages 97--101. Association for Computational Linguistics, San Diego,
  California, 2016.

\bibitem{shap}
Lundberg, S.~M., S.-I. Lee.
\newblock A unified approach to interpreting model predictions.
\newblock In I.~Guyon, U.~V. Luxburg, S.~Bengio, H.~Wallach, R.~Fergus,
  S.~Vishwanathan, R.~Garnett, eds., \emph{Advances in Neural Information
  Processing Systems 30}, pages 4765--4774. Curran Associates, Inc., 2017.

\bibitem{deeplift}
Shrikumar, A., P.~Greenside, A.~Kundaje.
\newblock Learning important features through propagating activation
  differences.
\newblock In D.~Precup, Y.~W. Teh, eds., \emph{Proceedings of the 34th
  International Conference on Machine Learning}, vol.~70 of \emph{Proceedings
  of Machine Learning Research}, pages 3145--3153. PMLR, 2017.

\bibitem{timeshap}
Bento, J., P.~Saleiro, A.~Cruz, et~al.
\newblock Timeshap: Explaining recurrent models through sequence perturbations.
\newblock In \emph{ACM Conference on Knowledge Discovery and Data Mining
  KDD'21}, pages~--. 2021.

\bibitem{deit}
Touvron, H., M.~Cord, M.~Douze, et~al.
\newblock Training data-efficient image transformers \& distillation through
  attention.
\newblock In M.~Meila, T.~Zhang, eds., \emph{Proceedings of the 38th
  International Conference on Machine Learning}, vol. 139 of \emph{Proceedings
  of Machine Learning Research}, pages 10347--10357. PMLR, 2021.

\bibitem{mtl_param_share1}
Pham, H., M.~Guan, B.~Zoph, et~al.
\newblock Efficient neural architecture search via parameters sharing.
\newblock In J.~Dy, A.~Krause, eds., \emph{Proceedings of the 35th
  International Conference on Machine Learning}, vol.~80 of \emph{Proceedings
  of Machine Learning Research}, pages 4095--4104. PMLR, 2018.

\bibitem{mtl_param_share2}
{Crawshaw}, M.
\newblock {Multi-Task Learning with Deep Neural Networks: A Survey}.
\newblock \emph{arXiv e-prints}, arXiv:2009.09796, 2020.

\bibitem{ts_kfold}
Bergmeir, C., R.~J. Hyndman, B.~Koo.
\newblock A note on the validity of cross-validation for evaluating
  autoregressive time series prediction.
\newblock \emph{Computational Statistics \& Data Analysis}, 120:70--83, 2018.

\bibitem{baq_dataset}
Zhang, S., B.~Guo, A.~Dong, et~al.
\newblock Cautionary tales on air-quality improvement in beijing.
\newblock \emph{Proceedings of the Royal Society A: Mathematical, Physical and
  Engineering Science}, 473:20170457, 2017.

\end{thebibliography}

\end{document}